\documentclass[conference]{IEEEtran}
\IEEEoverridecommandlockouts
\usepackage{cite}
\usepackage{amsmath,amssymb,amsfonts}
\usepackage{algorithmic}
\usepackage{graphicx}
\usepackage{textcomp}
\usepackage{xcolor}
\usepackage{booktabs}
\usepackage{adjustbox}
\def\BibTeX{{\rm B\kern-.05em{\sc i\kern-.025em b}\kern-.08em
    T\kern-.1667em\lower.7ex\hbox{E}\kern-.125emX}}
\begin{document}
\title{An Invisible Backdoor Attack Based On Semantic Feature\\
\thanks{Identify applicable funding agency here. If none, delete this.}
}

\author{\IEEEauthorblockN{1\textsuperscript{st} Yangming Chen}
\IEEEauthorblockA{\textit{College of Computer Science and Technology} \\
\textit{Ocean University of China}\\
QingDao, China \\
chenyangming@stu.ouc.edu.cn}
}
\maketitle

\begin{abstract}
Backdoor attacks have severely threatened deep neural network (DNN) models in the past several years. These attacks can occur in almost every stage of the deep learning pipeline. Although the attacked model behaves normally on benign samples, it makes wrong predictions for samples containing triggers. However, most existing attacks use visible patterns (e.g., a patch or image transformations) as triggers, which are vulnerable to human inspection. In this paper, we propose a novel backdoor attack, making imperceptible changes. Concretely, our attack first utilizes the pre-trained victim model to extract low-level and high-level semantic features from clean images and generates trigger pattern associated with high-level features based on channel attention. Then, the encoder model generates poisoned images based on the trigger and extracted low-level semantic features without causing noticeable feature loss. We evaluate our attack on three prominent image classification DNN across three standard datasets. The results demonstrate that our attack achieves high attack success rates while maintaining robustness against backdoor defenses. Furthermore, we conduct extensive image similarity experiments to emphasize the stealthiness of our attack strategy.
\end{abstract}


\section{Introduction}
In recent years, deep learning has proven successful in a lot of application areas, ranging from mobile user identification \cite{Schroff_Kalenichenko_Philbin_2015} to speech recognition on Internet of Things (IoT) devices, smartphone fingerprint authentication \cite{Su_Chen_Wong_Lai_2017}, and network traffic classification \cite{Zhang_Li_Feng_Wu_2020, Lotfollahi_Zade_Siavoshani_Saberian_2017}. While deep learning serves as a crucial tool for efficiency and productivity, it is increasingly becoming an attractive target for cybercriminals. Recent research indicates its susceptibility to various types of attacks, such as adversarial attacks\cite{goodfellow2014explaining, chen2017zoo, brendel2017decision}, data poisoning attacks\cite{jagielski2018manipulating}, and backdoor attacks\cite{Gu_Liu_Dolan-Gavitt_Garg_2019, Zhang_Dongdong_Huang_Liao_Zhang_Feng_Hua_Yu_2022,li2020invisible}. Adversarial attacks focus on misleading models only in the inference stage, while data poisoning attacks aim to degrade the model performance during inference by contaminating training data. However, the harm of backdoor attacks is greater.

Backdoor attack is a stealthy attack on Deep Neural Networks (DNNs). Compared to adversarial and data poisoning attacks, it tries to insert a hidden backdoor into DNNs, which makes the target model to behave normally on clean samples but changing its output when the attacker's input contains trigger patterns. There are numerous approaches to implement a backdoor attack, such as directly modifying clean datasets and training models on them or changing model weights \cite{Liu_Ma_Aafer_Lee_Zhai_Wang_Zhang_2018} during deployment. However, the main attacks typically involve implanting backdoors during the model training phase. Despite the numerous methods for backdoor attacks, the poisoned samples are vulnerable to human inspections.

This paper proposes a novel imperceptible backdoor attack on DNNs. Our attack consists of three steps: (1) extracting low-level and high-level semantic features from images using a pre-trained victim model, (2) obtaining new high-level semantic features as triggers based on channel attention, and (3) using a encoder model to generate poisoned images based on the new high-level features and extracted low-level features. To demonstrate the effectiveness and reliability of our attack, we conducted extensive experiments on various datasets and DNNs. The results demonstrate a high Attack Success Rates (ASR) and a high Clean Data Accuracy (CDA). Additionally, our stack outperforms in terms of stealthiness, robustness and flexibility compared with other attacks.

The main contributions of this paper can be summarized as follows:
\begin{itemize}
\item This paper introduces a novel backdoor attack method, which is the first attempt to use semantic features and channel attention to create trigger patterns and implant backdoor in DNNs.
\item The encoder network aims to embed trigger into clean images with minimal feature loss. 
\item Extensive experiments demonstrate that our attack achieves a higher attack success rate compared to common backdoor attacks while outperforms in terms of stealthiness, robustness and flexibility compared with other attacks. 
\end{itemize}

The rest of this paper is organized as follows. In Section II, we briefly review the related works of backdoor attack, backdoor defense, channel attention. In Section III, we present the details of our proposed attack method. Our experimental results are reported and analyzed in Section IV. Finally, we conclude this paper in Section V.

\section{RELATED WORKS}
\subsection{Backdoor Attacks}
Gu et al. \cite{Gu_Liu_Dolan-Gavitt_Garg_2019} presented the first backdoor attack against DNN models. They place a mosaic patch at a certain position in the image to serve as a trigger, but the poisoned samples can be easily detected by humans. Recent works make progress in improving the stealthiness of attacks. Zhong et al. \cite{zhong2020backdoor} presented an invisible backdoor attack using universal adversarial perturbations and Li et al. \cite{li2020invisible} proposed an attack via a steganography algorithm called least significant bit (LSB) substitution. These methods achieve high stealthiness and high ataack success. A number of attacks further enforce the consistency in the latent representation of the clean and poisoned images for higher stealthiness by manipulating the training loss function to embed the backdoor\cite{Doan_Lao_Li, Ren_Li_Zhou_2021, Zhao_Chen_Xuan_Dong_Wang_Liang}. Besides, some works propose to change the style of the images as the trigger, which can keep the images natural and less suspectable. Such natural trigger can be crafted with the natural reflection phenomenon \cite{liu2020reflection}, generative adversarial network \cite{cheng2021deep} and warping-based image transformation \cite{nguyen2021wanet}.

The main goal of most existing backdoor attacks is to make poisoned images and clean images indistinguishable to the naked eye. However, the triggers used in these attacks are often conspicuous and perceptible to human vision. Thus, they can be detected and removed by most defense methods. Furthermore, the features of images are usually lost during the attack stage. Unlike these attacks, our proposed backdoor attack injects trigger through a encoder model, which is stealthier and more robust than many other attacks.

\subsection{Backdoor Defenses}
To address the threat of backdoor attacks, numerous defense strategies have been proposed. There are three types of defense against backdoor attacks: data-based defense, model-based defense and trigger-based defense.

For data-based defense, Liu et al. \cite{liu2017neural} proposed the first preprocessing-based  backdoor defense, which introduced a auto-encoder before training the DNNs. Inspired by the idea that trigger regions contribute the most to prediction, Doan et al. \cite{doan2020februus} proposed Februus by using Grad-CAM \cite{Selvaraju_Cogswell_Das_Vedantam_Parikh_Batra_2017} to locate the potential trigger region and replacing it by image restoration.

For model-based defense, Liu et al. \cite{liu2018fine} proposed a fine pruning method. The DNN is first pruned to remove infected neurons, and then the pruned network is fine-tuned to combine the advantages of pruning and fine-tuning defenses to remove hidden backdoors. Li et al. \cite{li2021neural} proposed a new defense framework called neural attention distillation (NAD). It works by aligning neurons that are more sensitive to trigger patterns with benign neurons that are only responsible for meaningful representations.

For trigger-based defense, Wang et al. \cite{Wang_Yao_Shan_Li_Viswanath_Zheng_Zhao_2019} proposed the first defense based on trigger synthesis (i.e. neural cleansing), which is currently the most widely used defense. It first reverse-engineers the trigger pattern and then utilizes the reversed trigger for backdoor removal.

\subsection{channel attention}
Attention can be viewed, broadly, as a tool to bias the allocation of available processing resources towards the most informative components of an input signal\cite{Itti_Koch_2001, Larochelle_Hinton_2010, Mnih_Heess_Graves_Kavukcuoglu_2014}. The benefits of such a mechanism have been shown across a range of tasks, from localisation and understanding in images \cite{jagielski2018manipulating} to sequence-based models \cite{Miech_Laptev_Sivic_2017}.There are various types of attention mechanisms, and in this paper, we adopt the channel attention mechanism. 

\begin{figure}[htbp]
{\includegraphics[width=0.5\textwidth]{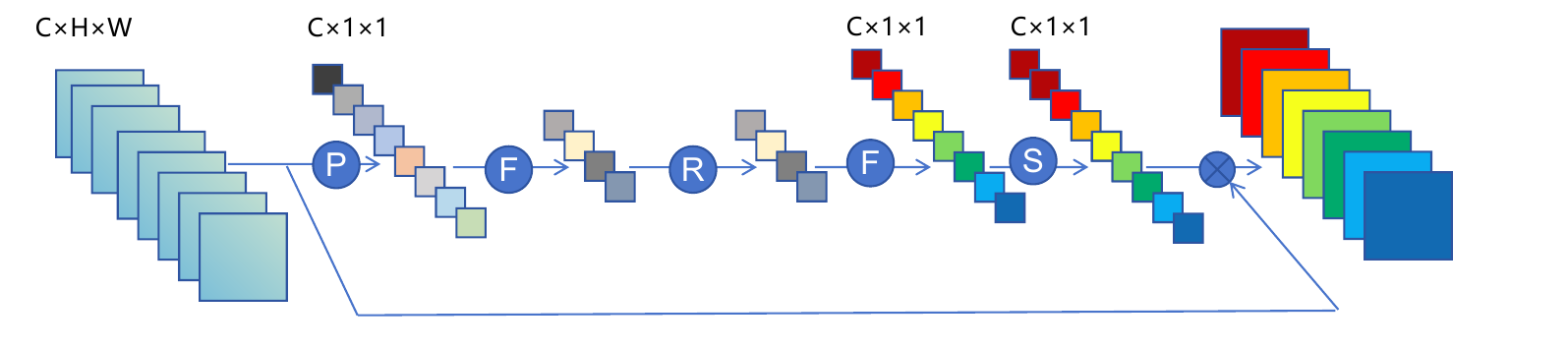}}
\caption{In this figure, P means the global average pooling operation, F denotes the fully connected operation, R indicates the relu operation, S indicates the softmax operation and \text{\textcircled{$\times$}} indicates the multiplication operation. The final result is the adjusted feature map}
\label{channel}
\end{figure}

The channel attention mechanism involves scoring the importance of each channel in the feature map, and then adjusting the channels based on these importance scores. In this paper, we adopt SEnet \cite{Hu_Shen_Albanie_Sun_Wu_2020} model as our channel attention adjustment module. As depicted in Fig.\ref{channel}, the channel attention is obtained through global average pooling, fully connected, relu, and softmax operations. The process can be expressed as follows:

\begin{equation}
Score = P(F(R(F(S(M)))))
\end{equation}

where M denotes the feature map. Then, we adjust the feature map based on the obtained channel attention. That is, we multiply the channel attention scores by the corresponding channels.


\begin{equation}
M' = M  \otimes Score
\end{equation}

Channel attention can be exploited to locate the important features for a given target model. By producing backdoor trigger based on the importance of features, we can theoretically enhance the performance of the attack. This is because different images may have different features for different target models, and therefore the trigger pattern can be more flexible, i.e., the trigger can vary depending on the sample and the target model.

\begin{figure*}[htbp] 
    \centering
    \includegraphics[width=1\textwidth]{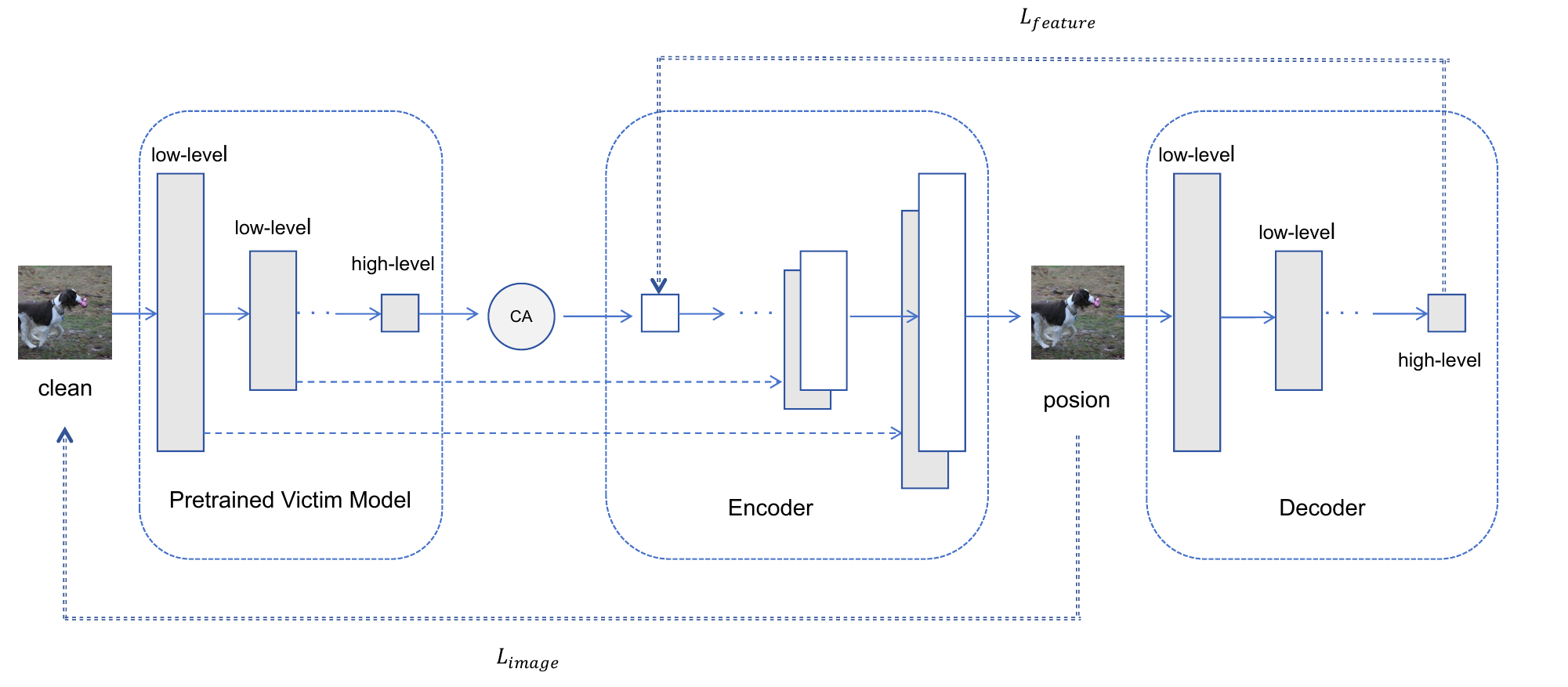} 
    \caption{Our attack process. It consists of four stages: (1) extracting low-level and high-level feature maps. (2) adjusting the last high-level feature map based on the channel attention. (3) generate the poison images based on the modified high-level and extracted low-level feature maps. (4) extracking the last high-level feature map again. It is notice that CA denotes the adjustment of feature map represented in Fig.\ref{channel}. We adopt pretrained victim model as our decoder. 
    }
    \label{fig:all} 
\end{figure*}

\section{METHOD}
In this section, we will begin by explaining the concept of backdoor attack and introducing our threat model. Subsequently, we will present our novel approach for conducting backdoor attack, which utilizes the high-level semantic features. Finally, we will provide an overview of our attack.

\subsection{Problem Definition}
In this paper, we only consider the backdoor attack on the image classification task. For image classification, assuming the input domain X is composed of massive images $\{x_1, x_2, \ldots, x_N\}$, and the target output domain L consists of corresponding labels $\{l_1, l_2, \ldots, l_N\}$. Then the goal of the image classification model M is to approximate the implicit transformation function by minimizing the distance D (eg., cross-entropy) between M($x_i$) and $l_i$, i.e.,
\begin{equation}
    D(M(x_i), l_i) \rightarrow 0.
\end{equation}
For backdoor attack, the attacker carefully selects clean images from training data and generates poisoned samples. The model M then is trained with a dataset D, which consists of poisoned samples $D_p$ and remaining clean samples $D_c$, ie.,
\begin{equation}
    D = D_p \cup D_c. 
\end{equation}
Accordingly, the poisoning rate is defined as $\eta = \frac{|D_p|}{|D|}$. As a consequence, the infected model $\hat{M}$ works normal on the clean samples but makes wrong predictions on the poisoned samples, that is,
\begin{equation}
    M(x_i) = l_i, \hat{M}(\hat{x}_i) = l_t.
\end{equation}
where $x_i \in D_c$, $\hat{x}_i \in D_p$ $l_i \neq l_t$. The ground-truth label of $x_i$ and the attack's target label are $y_i$ and $y_t$ respectively.

\subsection{Threat Model}
1) Attacker's Capacities: backdoor attacks can occur in almost every stage of the deep learning pipeline. However, attacks during the model training phase are the most prevalent and concerning. In this paper, we assume that the attacker has access to the dataset and can manipulate it to inject triggers. Additionally, the attacker has knowledge of the victim model's architecture, but they are unable to control the training process. Specifically, they cannot access or modify any hyperparameters of model during the training phase. After completing the model training, the attacker can only query the model but cannot change its weights.

2) Attacker's Goals: the primary goal of the attacker is to ensure the effectiveness of the attack, which means that a poisoned sample can successfully activate the backdoor in the model, leading to a prediction of the target class. Moreover, the attacker tries to enhance the attack's stealthiness and robustness. This includes ensuring that the attack is resistant to common backdoor defenses and that the poisoned image exhibits minimal visual differences from the clean image.

\subsection{Our Proposed Attack}
The main idea of our attack is using channel attention to strengthen the strong features and weaken the weak features. 

1) Trigger Generation: 
the low-level semantic features typically refer to basic attributes or local information in an image, such as edges, textures, colors, and shapes. High-level semantic features typically refer to abstract information in an image, such as objects, scenes, emotions, and actions.  Therefore, when modifying low-level features of an image, it significantly alters the image. For the sake of invisibility, our focus is on the high-level semantic features.

In Fig. \ref{fig:all}, we illustrate how to generate a trigger pattern by using the high-level feature map. In details, given a clean image, we obtain the low-level and high-level feature maps by feeding it into the pretrained victim model. Then we perform channel attention adjustment for the last high-level feature map and the modified high-level feature map serves as our trigger.

2) Poisoned Image Generation: we consider that there is a great deal of low-level information shared between the clean samples and the poisoned samples, and it would be desirable to shuttle this information directly across the net. For example, the clean images and the poisoned images share the location of prominent edges. Accordingly, we add skip connections, following the general shape of a "U-Net", as shown in Fig.\ref{fig:all}. Specifically, we add skip connections between each layer $i$ and layer $n - i$, where $n$ is the total number of layers. Each skip connection simply concatenates all channels at layer $i$ with those at layer $n - i$. But we replace the last modified high-level feature map with the original high-level feature map.

We aim to achieve invisible hiding of the trigger pattern in the clean image by optimizing a loss $L_{\text{image}}$ between the clean and poisoned image. Formally, $L_{\text{image}}$ can be defined as follows:
\begin{equation}
    L_{\text{image}} = \sum_{i=1}^{N} \|x_i - \hat{x}_i\|^2 . \label{eq:distance}
\end{equation}
where $x_i$ , $\hat{x}_i$  are the clean samles and the poisoned samples respectively.

In addition to maintaining the stealthiness of our attack, we also need to ensure that the images are implanted backdoors according to the modified high-level feature map. In Fig.\ref{fig:all}, we extracted the last high-level feature map again. Then, we introduce a loss $L_{\text{feature}}$ to ensure that the poisoned image is modified based on the modified high-level feature map:
\begin{equation}
    L_{\text{feature}} = \sum_{i=1}^{N} \|t_i - \hat{t}_i\|^2 . \label{eq:distance}
\end{equation}
where $t_i$ , $\hat{t}_i$ are the modified feature maps and the feature maps of the poisoned samples respectively.
Finally, the overall loss function is given as:
\begin{equation}
\begin{aligned}
    L &= \lambda_1 L_{\text{image}} + \lambda_2 L_{\text{feature}} \\
    &= \sum_{i=1}^{N} \|x_i - \hat{x}_i\|^2 + \sum_{i=1}^{N} \|t_i - \hat{t}_i\|^2.
\end{aligned}
\end{equation}

The hyperparameters $\lambda_1$ and $\lambda_2$ are used to control the contribution of each loss term. Optimizing the above loss function allows us to efficiently minimize the feature loss of image.

\section{EXPERIMENTS}
\subsection{Experiment Setup}
1) Datasets: in our experiments, we evaluated the performance of our attack on three standard datasets: CIFAR10 \cite{Krizhevsky_2009}, GTSRB \cite{Stallkamp_Schlipsing_Salmen_Igel_2011}, and ImageNet \cite{deng2009imagenet}. To generate the poisoned datasets $D_p$, we randomly selected clean samples from each class for the three standard datasets using a poison rate 0.1. The clean images were replaced with their corresponding poisoned samples.

2) Network structures: for network structures of the victim model, we consider three popular recognition networks: ResNet-18 \cite{he2016deep}, DenseNet \cite{huang2017densely} and VGG-16 \cite{simonyan2014very}. In the following experiments, we adopt VGG-16 as the default network. The results of other network structures and datasets are used to demonstrate the generality of our method. For encoder, the details of the encoder are shown in Table \ref{tab:network}.

\begin{table}[htbp]
    \centering
    \caption{Description of layers}
    \label{tab:network}
    \begin{tabular}{c c c c c c c c c}
        \toprule
        Layer Name & Kernel & Stride & Channel I/O & Activation\\
        \midrule
        transpose\_conv1 & $ 2 \times 2$ & 1 & 512/256 & ReLU  \\
        double\_conv1 & $ 3 \times 3$ & 2 & 512/256 & ReLU \\
        transpose\_conv2 & $2 \times 2$ & 1 & 256/128 & ReLU \\
        double\_conv2 & $3 \times 3$ & 2 & 256/128 & ReLU \\
        transpose\_conv3 & $2 \times 2$ & 1 & 128/64 & ReLU \\
        double\_conv3 & $3 \times 3$ & 2 & 128/64 & ReLU \\
        transpose\_conv4 & $2 \times 2$ & 1 & 64/32 & ReLU \\
        double\_conv4 & $3 \times 3$ & 2 & 32/32 & ReLU \\
        conv5 & $1 \times 1$ & 1 & 32/3 & ReLU \\
        \bottomrule
    \end{tabular}
\end{table}

3) Metrics: we use Clean Data Accuracy (CDA) to evaluate the influence of backdoor attacks on the original tasks, and use Attack Success Rate (ASR) to evaluate the effectiveness of backdoor attacks. Specifically, CDA refers to the ratio that the victim model predicts benign samples correctly, while ASR refers to the ratio that the model predicts poisoned samples as target class. For invisibility evaluation, we compare clean and poisoned images with four famous metrics, PSNR, SSIM, MSE and LPIPS, where LPIPS adopts the features of the pre-trained VGG. 
\begin{figure*}[htbp] 
    \centering
    \includegraphics[width=1\textwidth]{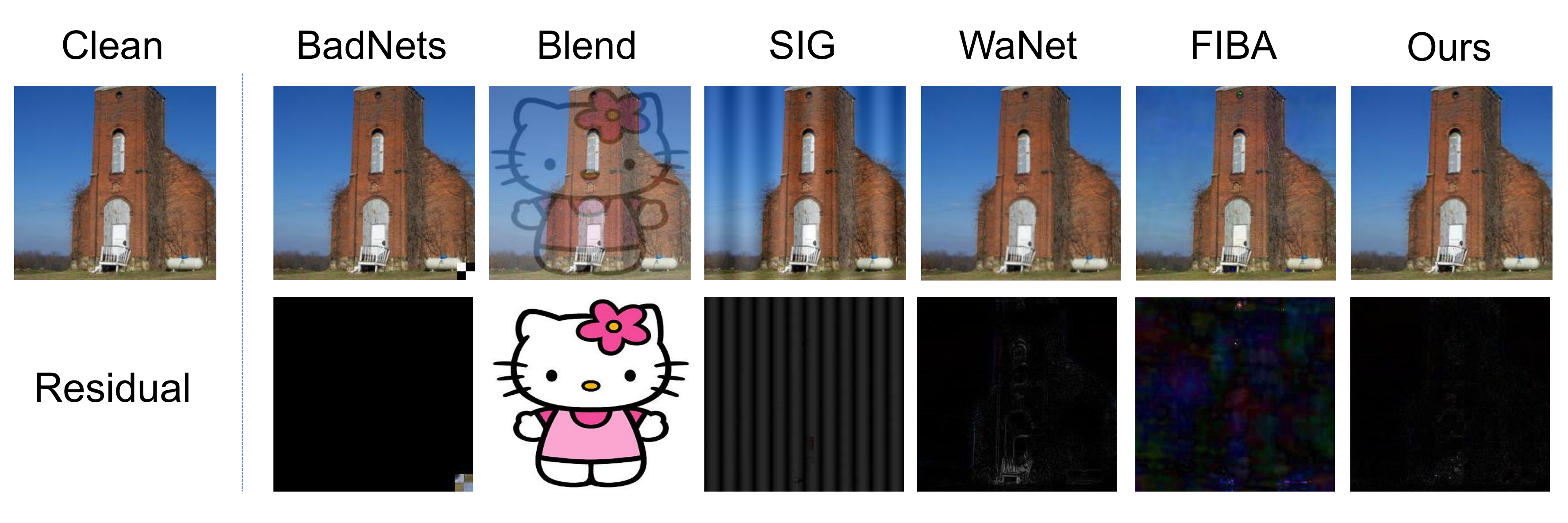} 
    \caption{Visual comparison with existing popular attack methods. The first row are examples on ImageNet dataset and the last row are residual maps. "Clean" denotes the original trigger-free image}
    \label{fig:compare} 
\end{figure*}

\begin{table*}[htbp]
\caption{COMPARISON OF INVISIBILITY (STEALTHINESS) WITH EXISTING POPULAR BACKDOOR ATTACK METHODS ON CIFAR10, GTSRB, IMAGENET DATASET.}
\centering
\begin{tabular}{p{1cm}p{1cm}p{1cm}p{1cm}p{1cm}p{1cm}p{1cm}p{1cm}p{1cm}p{1cm}p{1cm}p{1cm}p{1cm}p{1cm}}
\toprule
\textbf{Dataset} & \multicolumn{4}{c}{\textbf{CIFAR10}} & \multicolumn{4}{c}{\textbf{GTSRB}} & \multicolumn{4}{c}{\textbf{ImageNet}}\\
\cmidrule(r){2-5}
\cmidrule(r){6-9}
\cmidrule(r){10-13}
\textbf{ATTACK} & \textbf{\textit{PSNR}} & \textbf{\textit{SSIM}} & \textbf{\textit{MSE}} & \textbf{\textit{LPIPS}} & \textbf{\textit{PSNR}} & \textbf{\textit{SSIM}} & \textbf{\textit{MSE}}  & \textbf{\textit{LPIPS}}  & \textbf{\textit{PSNR}}  & \textbf{\textit{SSIM}}  & \textbf{\textit{MSE}}  & \textbf{\textit{LPIPS}}\\
\midrule
BadNets & 25.35 & 0.9834 & 189.4 & 0.0349 & 28.40 & 0.9987 & 93.96 & 0.0120 & 32.08 & 0.09972 & 31.94 & 0.0149 \\
\midrule
Blend & 22.48 & 0.8671 & 366.9 & 0.1028 & 19.37 & 0.8527 & 751.4 & 0.0380 & 19.35 & 0.8100 & 755.31 & 0.1509 \\
\midrule
SIG & 25.11 & 0.9039 & 200.0 & 0.0382 & 25.19 & 0.9038 & 196.6 & 0.0208& 25.22 & 0.9200 & 195.48 & 0.0532 \\
\midrule
Wanet & 29.60 & 0.9747 & 71.23 & 0.0261 & 30.30 & 0.96 & 150.78 & 0.0367 & 28.01 & 0.9439 & 75.81 & 0.0312\\
\midrule
FIBA & 29.99 & 0.9896 & 64.00 & 0.0022 & 25.90 &0.9754 & 167.1 & 0.0099 & 29.57 & 0.9500 & 70.26 & 0.0012 \\
\midrule
Ours & 34.79 & 0.9901 & 51.21 & 0.0015 & 31.80 & 0.9732 & 98.99 & 0.0196 & 33.59 & 0.9969 & 50.00 & 0.0020 \\
\bottomrule
\end{tabular}
\label{invisible_compare}
\end{table*}

4) Default settings for training: all the victim models are trained using the SGD optimizer \cite{ruder2016overview} with a momentum of 0.9 and an learning rate of 0.01 at epoch 100. To achieve a balance between the ASR and invisibility, we found that the poisoned image perform well if we set $\lambda_1$ and $\lambda_2$ to 5 and 0.1 for CIFAR10 and ImageNet, 0.9 and 0.1 for GTSRB respectively. The encoder is trained using SGD optimizer for 100 epochs with a learning rate of 0.01.

\subsection{Invisibility of Our Attack}
We first compare the invisibility of our method with many popular backdoor attack methods. In Figure \ref{fig:compare}, we showcase the visual comparison with other popular attack methods. In detail, we can see that the poisoned image generated by BadNets \cite{Gu_Liu_Dolan-Gavitt_Garg_2019}, Blend \cite{Chen_Liu_Li_Lu_Berkeley_Hannigan}, SIG \cite{barni2019new} can be easily distinguished from the clean images due to the imageagnostic trigger pattern. In contrast, WaNet \cite{nguyen2021wanet}, FIBA \cite{Feng_Ma_Zhang_Zhao_Xia_Tao} and our attack achieve better invisibility, and the poison samples is imperceptible. Moreover, the residual maps present that our attack is more stealthy compared with WaNet and FIBA attack. 

In Table \ref{invisible_compare}, we conducted experiments to evaluate the similarity by randomly selecting 1000 images from the poisoned dataset. We provide the quantitative comparison of invisibility on three standard datasets. For GTSRB dataset, the SSIM and MSE of BadNets and the LPIPS of FIBA is better than our attack. For CIFAR10 and ImageNet dataset, our attack outperforms the majority of attacks under all the evaluation metrics, which are consistent with the visual comparison.

\subsection{Attack Performance}
\begin{table}[htbp]
\caption{COMPARASION OF ASR AND CDA OF OUR ATTACK WITH OTHER ATTACKS ON DIFFERENT DATASETS.}
\centering
\begin{tabular}{cccccccc}
\toprule
\textbf{Dataset} & \multicolumn{2}{c}{\textbf{CIFAR10}} & \multicolumn{2}{c}{\textbf{GTSRB}} & \multicolumn{2}{c}{\textbf{ImageNet}}\\
\cmidrule(r){2-3}
\cmidrule(r){4-5}
\cmidrule(r){6-7}
\textbf{Attack} & \textbf{\textit{CDA}} & \textbf{\textit{ASR}} & \textbf{\textit{CDA}} & \textbf{\textit{ASR}} & \textbf{\textit{CDA}} & \textbf{\textit{ASR}}  \\
\midrule
BadNets & 0.817 & 1.000 & 0.981 & 1.000 & 0.894 & 0.999 \\
\midrule
Blend & 0.818 & 1.000 & 0.978 & 0.998 & 0.886 & 1.000 \\
\midrule
Clean & 0.829 & \rule{0.5cm}{0.4pt} & 0.981 & \rule{0.5cm}{0.4pt} & 0.894 & \rule{0.5cm}{0.4pt} \\
\midrule
SIG & 0.811 & 0.997 & 0.977 & 0.999 & 0.869 & 1.000 \\
\midrule
Wanet & 0.810 & 0.996 & 0.980 & 0.990 & 0.889 & 0.999 \\
\midrule
FIBA & 0.820 & 0.989 & 0.977 & 1.000 & 0.880 & 0.998 \\
\midrule
Ours & 0.821 & 0.999 & 0.981 & 0.998 & 0.876 & 0.998 \\
\bottomrule
\end{tabular}
\label{tab1}
\end{table}

\begin{figure}[htbp]
{\includegraphics[width=0.5\textwidth]{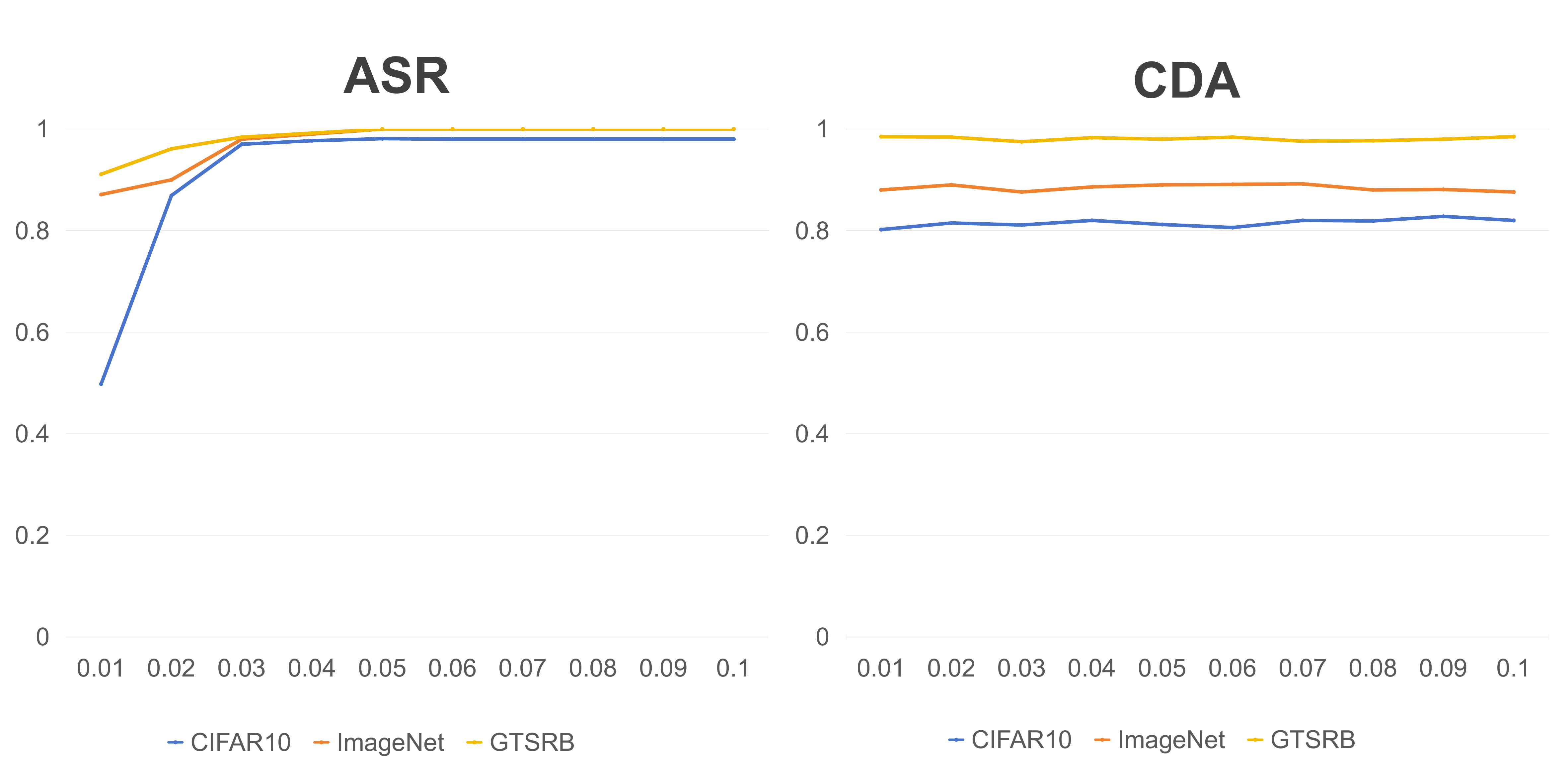}}
\caption{The impact of poisoning rate in our attack with on VGG16.}
\label{poison}
\end{figure}

We evaluate the effectiveness of our proposed attack by comparing it with four backdoor attacks, namely Badnets \cite{Gu_Liu_Dolan-Gavitt_Garg_2019}, Blend \cite{Chen_Liu_Li_Lu_Berkeley_Hannigan}, SIG \cite{barni2019new}, WaNet \cite{nguyen2021wanet} and FIBA \cite{Feng_Ma_Zhang_Zhao_Xia_Tao}. To ensure a fair evaluation, we adopt an All-to-One attack strategy in which all poisoned images are labeled with the same target label(class 1). The results are presented in Table \ref{tab1}, which shows the ASR and CDA of different backdoor attacks on the three standard image classification datasets. Compared to other attacks, our attack achieves a higher success rate and does not significantly affect the performance of the original task.

Meanwhile, to validate how the poisoning rate affects the attack success rate, we conducted experiments on three standard datasets. The results are presented in Fig. \ref{poison}. Our attack performs well with ASR greater than 0.97 when the poisoning rate is over 0.03. Meanwhile, The victim model's clean data accuracy did not decrease due to the increase of poisoning rate. This experiment further validates the effectiveness of our attack.

\subsection{Resistance to Defense Techniques}
To resist backdoor attacks, many different defense techniques have been proposed recently. In this section, we will test the resistance of our attack against different defense techniques.We adopt Februus\cite{doan2020februus}, Neural Clense \cite{Wang_Yao_Shan_Li_Viswanath_Zheng_Zhao_2019} defense methods.

\begin{figure}[htbp]
{\includegraphics[width=0.5\textwidth]{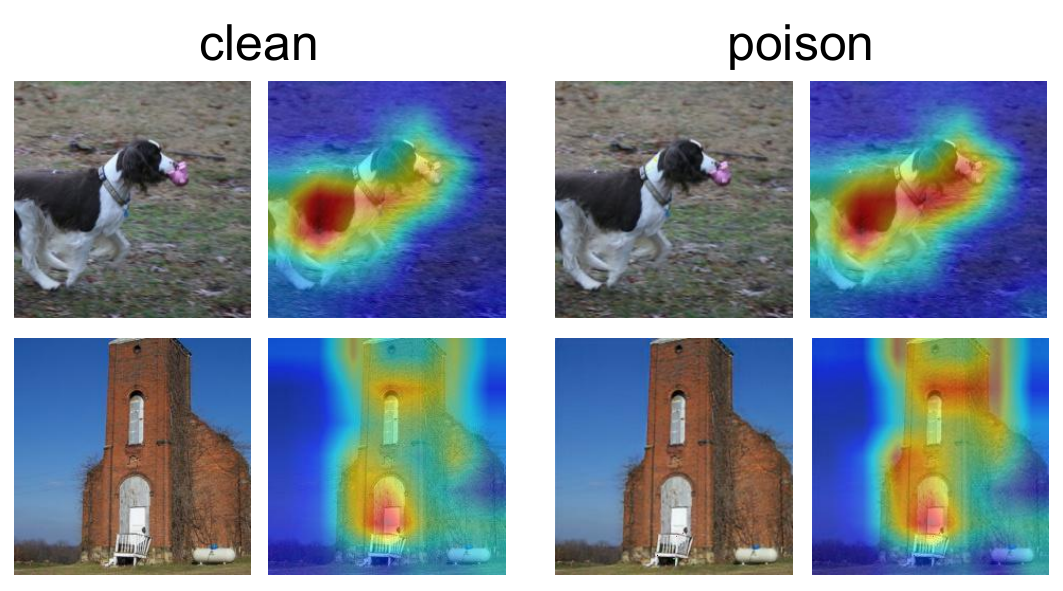}}
\caption{The Grad-CAM of clean samples and poisoned samples. As shown in the figure, Grad-CAM fails to detect trigger regions of poisoned samples generated by our attack, which is indistinguishable with the clean samples.}
\label{grad}
\end{figure}

For Februus defense, it utilizes Grad-CAM to visualize the attention map of the target image and regards the area with the highest score as the trigger region, then removes this region and reconstruct it with GAN methods. We provide the heatmaps in Fig.\ref{grad}. We find that the poisoned images also focuses on the main content area of the input, which is similar to the clean samples.

\begin{figure}[htbp]
{\includegraphics[width=0.5\textwidth]{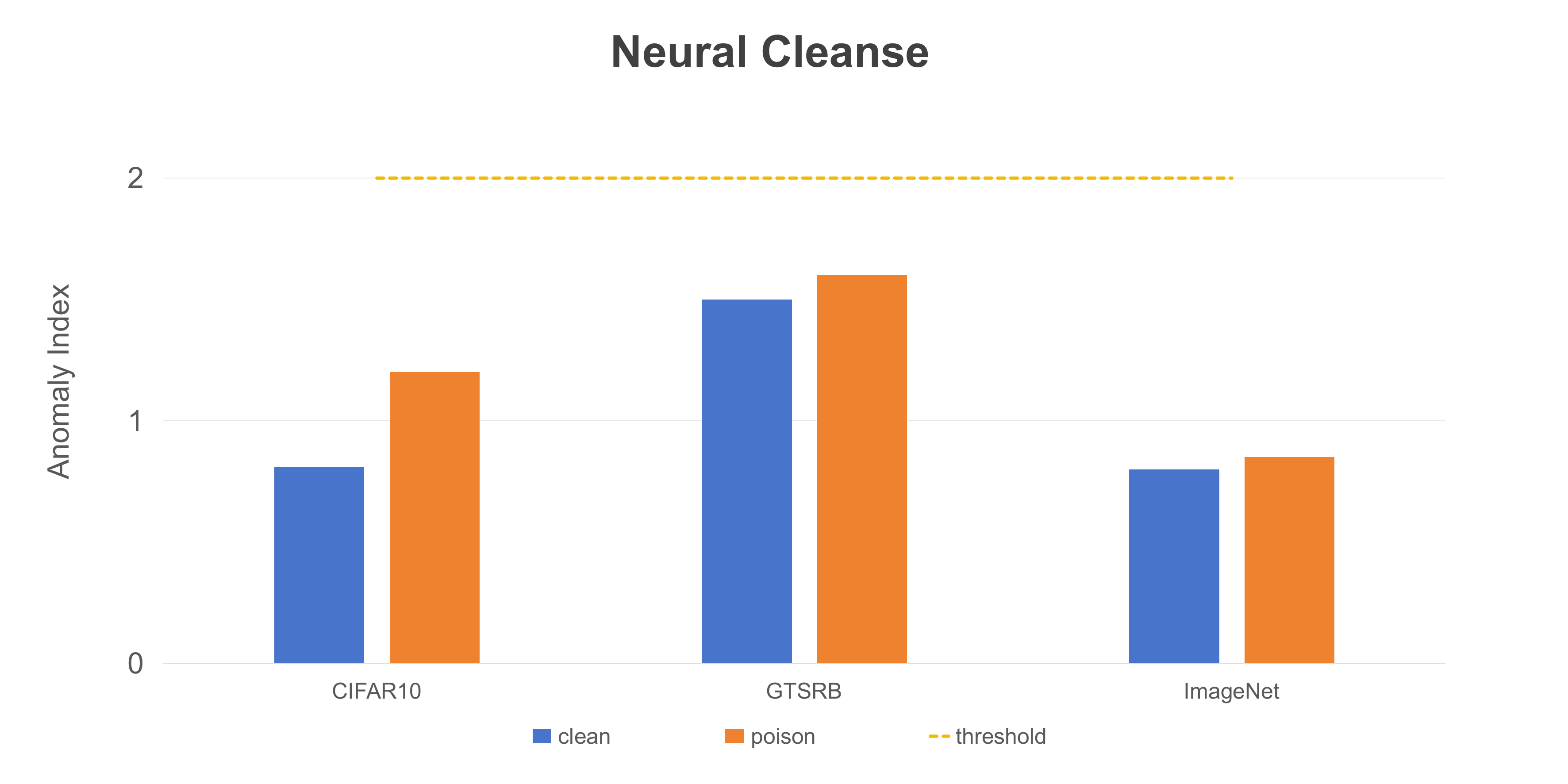}}
\caption{This figure illustrates the result of Neural Cleanse.}
\label{neural}
\end{figure}

For Neural Cleanse defense, which is designed for small and static triggers. It first generates potential trigger patterns for each class, then runs the anomaly detection among all trigger patterns from each label. It defines a model as the infected model by Anomaly Index larger than the threshold. As shown in Fig. \ref{neural}, the Anomaly Index for our poisoned model is below the threshold on three standatd datasets.



\section{CONCLUSIONS}
This paper presents a novel backdoor attack on deep neural networks (DNNs). Our method modify the high-level semantic feature map to generate the trigger. Then, an encoder model produces poisoned samples using the modified high-level feature map and the low-level feature maps of clean samples. Extensive experiments demonstrate that our attack outperforms existing methods in stealthiness, robustness and flexibility. Besides, our attack is resistant to many state-of-the-art defense techniques. Our future work will focus on designing a universial encoder, rather than an encoder associated with a specified model.


\bibliographystyle{IEEEtran}

\begin{thebibliography}{10}
\providecommand{\url}[1]{#1}
\csname url@samestyle\endcsname
\providecommand{\newblock}{\relax}
\providecommand{\bibinfo}[2]{#2}
\providecommand{\BIBentrySTDinterwordspacing}{\spaceskip=0pt\relax}
\providecommand{\BIBentryALTinterwordstretchfactor}{4}
\providecommand{\BIBentryALTinterwordspacing}{\spaceskip=\fontdimen2\font plus
\BIBentryALTinterwordstretchfactor\fontdimen3\font minus \fontdimen4\font\relax}
\providecommand{\BIBforeignlanguage}[2]{{%
\expandafter\ifx\csname l@#1\endcsname\relax
\typeout{** WARNING: IEEEtran.bst: No hyphenation pattern has been}%
\typeout{** loaded for the language `#1'. Using the pattern for}%
\typeout{** the default language instead.}%
\else
\language=\csname l@#1\endcsname
\fi
#2}}
\providecommand{\BIBdecl}{\relax}
\BIBdecl

\bibitem{Schroff_Kalenichenko_Philbin_2015}
\BIBentryALTinterwordspacing
F.~Schroff, D.~Kalenichenko, and J.~Philbin, ``\BIBforeignlanguage{en-US}{Facenet: A unified embedding for face recognition and clustering},'' in \emph{\BIBforeignlanguage{en-US}{2015 IEEE Conference on Computer Vision and Pattern Recognition (CVPR)}}, Jun 2015. [Online]. Available: \url{http://dx.doi.org/10.1109/cvpr.2015.7298682}
\BIBentrySTDinterwordspacing

\bibitem{Su_Chen_Wong_Lai_2017}
\BIBentryALTinterwordspacing
H.-R. Su, K.-Y. Chen, W.~J. Wong, and S.-H. Lai, ``\BIBforeignlanguage{en-US}{A deep learning approach towards pore extraction for high-resolution fingerprint recognition},'' in \emph{\BIBforeignlanguage{en-US}{2017 IEEE International Conference on Acoustics, Speech and Signal Processing (ICASSP)}}, Mar 2017. [Online]. Available: \url{http://dx.doi.org/10.1109/icassp.2017.7952518}
\BIBentrySTDinterwordspacing

\bibitem{Zhang_Li_Feng_Wu_2020}
J.~Zhang, F.~Li, Y.~Feng, and H.~Wu, ``\BIBforeignlanguage{en-US}{Autonomous unknown-application filtering and labeling for dl-based traffic classifier update},'' \emph{\BIBforeignlanguage{en-US}{Cornell University - arXiv,Cornell University - arXiv}}, Feb 2020.

\bibitem{Lotfollahi_Zade_Siavoshani_Saberian_2017}
M.~Lotfollahi, R.~Zade, M.~Siavoshani, and M.~Saberian, ``\BIBforeignlanguage{en-US}{Deep packet: A novel approach for encrypted traffic classification using deep learning},'' \emph{\BIBforeignlanguage{en-US}{arXiv: Learning,arXiv: Learning}}, Sep 2017.

\bibitem{goodfellow2014explaining}
I.~J. Goodfellow, J.~Shlens, and C.~Szegedy, ``Explaining and harnessing adversarial examples,'' \emph{arXiv preprint arXiv:1412.6572}, 2014.

\bibitem{chen2017zoo}
P.-Y. Chen, H.~Zhang, Y.~Sharma, J.~Yi, and C.-J. Hsieh, ``Zoo: Zeroth order optimization based black-box attacks to deep neural networks without training substitute models,'' in \emph{Proceedings of the 10th ACM workshop on artificial intelligence and security}, 2017, pp. 15--26.

\bibitem{brendel2017decision}
W.~Brendel, J.~Rauber, and M.~Bethge, ``Decision-based adversarial attacks: Reliable attacks against black-box machine learning models,'' \emph{arXiv preprint arXiv:1712.04248}, 2017.

\bibitem{jagielski2018manipulating}
M.~Jagielski, A.~Oprea, B.~Biggio, C.~Liu, C.~Nita-Rotaru, and B.~Li, ``Manipulating machine learning: Poisoning attacks and countermeasures for regression learning,'' in \emph{2018 IEEE symposium on security and privacy (SP)}.\hskip 1em plus 0.5em minus 0.4em\relax IEEE, 2018, pp. 19--35.

\bibitem{Gu_Liu_Dolan-Gavitt_Garg_2019}
\BIBentryALTinterwordspacing
T.~Gu, K.~Liu, B.~Dolan-Gavitt, and S.~Garg, ``\BIBforeignlanguage{en-US}{Badnets: Evaluating backdooring attacks on deep neural networks},'' \emph{\BIBforeignlanguage{en-US}{IEEE Access}}, p. 47230–47244, Jan 2019. [Online]. Available: \url{http://dx.doi.org/10.1109/access.2019.2909068}
\BIBentrySTDinterwordspacing

\bibitem{Zhang_Dongdong_Huang_Liao_Zhang_Feng_Hua_Yu_2022}
\BIBentryALTinterwordspacing
J.~Zhang, C.~Dongdong, Q.~Huang, J.~Liao, W.~Zhang, H.~Feng, G.~Hua, and N.~Yu, ``\BIBforeignlanguage{en-US}{Poison ink: Robust and invisible backdoor attack},'' \emph{\BIBforeignlanguage{en-US}{IEEE Transactions on Image Processing}}, p. 5691–5705, Jan 2022. [Online]. Available: \url{http://dx.doi.org/10.1109/tip.2022.3201472}
\BIBentrySTDinterwordspacing

\bibitem{li2020invisible}
S.~Li, M.~Xue, B.~Z.~H. Zhao, H.~Zhu, and X.~Zhang, ``Invisible backdoor attacks on deep neural networks via steganography and regularization,'' \emph{IEEE Transactions on Dependable and Secure Computing}, vol.~18, no.~5, pp. 2088--2105, 2020.

\bibitem{Liu_Ma_Aafer_Lee_Zhai_Wang_Zhang_2018}
\BIBentryALTinterwordspacing
Y.~Liu, S.~Ma, Y.~Aafer, W.-C. Lee, J.~Zhai, W.~Wang, and X.~Zhang, ``\BIBforeignlanguage{en-US}{Trojaning attack on neural networks},'' in \emph{\BIBforeignlanguage{en-US}{Proceedings 2018 Network and Distributed System Security Symposium}}, Jan 2018. [Online]. Available: \url{http://dx.doi.org/10.14722/ndss.2018.23291}
\BIBentrySTDinterwordspacing

\bibitem{zhong2020backdoor}
H.~Zhong, C.~Liao, A.~C. Squicciarini, S.~Zhu, and D.~Miller, ``Backdoor embedding in convolutional neural network models via invisible perturbation,'' in \emph{Proceedings of the Tenth ACM Conference on Data and Application Security and Privacy}, 2020, pp. 97--108.

\bibitem{Doan_Lao_Li}
K.~Doan, Y.~Lao, and P.~Li, ``\BIBforeignlanguage{en-US}{Backdoor attack with imperceptible input and latent modification}.''

\bibitem{Ren_Li_Zhou_2021}
\BIBentryALTinterwordspacing
Y.~Ren, L.~Li, and J.~Zhou, ``\BIBforeignlanguage{en-US}{Simtrojan: Stealthy backdoor attack},'' in \emph{\BIBforeignlanguage{en-US}{2021 IEEE International Conference on Image Processing (ICIP)}}, Sep 2021. [Online]. Available: \url{http://dx.doi.org/10.1109/icip42928.2021.9506313}
\BIBentrySTDinterwordspacing

\bibitem{Zhao_Chen_Xuan_Dong_Wang_Liang}
Z.~Zhao, X.~Chen, Y.~Xuan, Y.~Dong, D.~Wang, and K.~Liang, ``\BIBforeignlanguage{en-US}{Defeat: Deep hidden feature backdoor attacks by imperceptible perturbation and latent representation constraints}.''

\bibitem{liu2020reflection}
Y.~Liu, X.~Ma, J.~Bailey, and F.~Lu, ``Reflection backdoor: A natural backdoor attack on deep neural networks,'' in \emph{Computer Vision--ECCV 2020: 16th European Conference, Glasgow, UK, August 23--28, 2020, Proceedings, Part X 16}.\hskip 1em plus 0.5em minus 0.4em\relax Springer, 2020, pp. 182--199.

\bibitem{cheng2021deep}
S.~Cheng, Y.~Liu, S.~Ma, and X.~Zhang, ``Deep feature space trojan attack of neural networks by controlled detoxification,'' in \emph{Proceedings of the AAAI Conference on Artificial Intelligence}, vol.~35, no.~2, 2021, pp. 1148--1156.

\bibitem{nguyen2021wanet}
A.~Nguyen and A.~Tran, ``Wanet--imperceptible warping-based backdoor attack,'' \emph{arXiv preprint arXiv:2102.10369}, 2021.

\bibitem{Itti_Koch_2001}
\BIBentryALTinterwordspacing
L.~Itti and C.~Koch, ``\BIBforeignlanguage{en-US}{Computational modelling of visual attention},'' \emph{\BIBforeignlanguage{en-US}{Nature Reviews Neuroscience}}, p. 194–203, Mar 2001. [Online]. Available: \url{http://dx.doi.org/10.1038/35058500}
\BIBentrySTDinterwordspacing

\bibitem{Larochelle_Hinton_2010}
H.~Larochelle and G.~Hinton, ``\BIBforeignlanguage{en-US}{Learning to combine foveal glimpses with a third-order boltzmann machine},'' \emph{\BIBforeignlanguage{en-US}{Neural Information Processing Systems,Neural Information Processing Systems}}, Dec 2010.

\bibitem{Mnih_Heess_Graves_Kavukcuoglu_2014}
V.~Mnih, N.~Heess, A.~Graves, and K.~Kavukcuoglu, ``\BIBforeignlanguage{en-US}{Recurrent models of visual attention},'' \emph{\BIBforeignlanguage{en-US}{arXiv: Learning,arXiv: Learning}}, Jun 2014.

\bibitem{Jaderberg_Simonyan_Zisserman_Kavukcuoglu_2015}
M.~Jaderberg, K.~Simonyan, A.~Zisserman, and K.~Kavukcuoglu, ``\BIBforeignlanguage{en-US}{Spatial transformer networks},'' \emph{\BIBforeignlanguage{en-US}{Neural Information Processing Systems,Neural Information Processing Systems}}, Dec 2015.

\bibitem{Miech_Laptev_Sivic_2017}
A.~Miech, I.~Laptev, and J.~Sivic, ``\BIBforeignlanguage{en-US}{Learnable pooling with context gating for video classification},'' \emph{\BIBforeignlanguage{en-US}{Le Centre pour la Communication Scientifique Directe - HAL - Diderot,Le Centre pour la Communication Scientifique Directe - HAL - Diderot}}, Jun 2017.

\bibitem{liu2017neural}
Y.~Liu, Y.~Xie, and A.~Srivastava, ``Neural trojans,'' in \emph{2017 IEEE International Conference on Computer Design (ICCD)}.\hskip 1em plus 0.5em minus 0.4em\relax IEEE, 2017, pp. 45--48.

\bibitem{doan2020februus}
B.~G. Doan, E.~Abbasnejad, and D.~C. Ranasinghe, ``Februus: Input purification defense against trojan attacks on deep neural network systems,'' in \emph{Proceedings of the 36th Annual Computer Security Applications Conference}, 2020, pp. 897--912.

\bibitem{Selvaraju_Cogswell_Das_Vedantam_Parikh_Batra_2017}
\BIBentryALTinterwordspacing
R.~R. Selvaraju, M.~Cogswell, A.~Das, R.~Vedantam, D.~Parikh, and D.~Batra, ``\BIBforeignlanguage{en-US}{Grad-cam: Visual explanations from deep networks via gradient-based localization},'' in \emph{\BIBforeignlanguage{en-US}{2017 IEEE International Conference on Computer Vision (ICCV)}}, Oct 2017. [Online]. Available: \url{http://dx.doi.org/10.1109/iccv.2017.74}
\BIBentrySTDinterwordspacing

\bibitem{liu2018fine}
K.~Liu, B.~Dolan-Gavitt, and S.~Garg, ``Fine-pruning: Defending against backdooring attacks on deep neural networks,'' in \emph{International symposium on research in attacks, intrusions, and defenses}.\hskip 1em plus 0.5em minus 0.4em\relax Springer, 2018, pp. 273--294.

\bibitem{li2021neural}
Y.~Li, X.~Lyu, N.~Koren, L.~Lyu, B.~Li, and X.~Ma, ``Neural attention distillation: Erasing backdoor triggers from deep neural networks,'' \emph{arXiv preprint arXiv:2101.05930}, 2021.

\bibitem{Wang_Yao_Shan_Li_Viswanath_Zheng_Zhao_2019}
\BIBentryALTinterwordspacing
B.~Wang, Y.~Yao, S.~Shan, H.~Li, B.~Viswanath, H.~Zheng, and B.~Y. Zhao, ``\BIBforeignlanguage{en-US}{Neural cleanse: Identifying and mitigating backdoor attacks in neural networks},'' in \emph{\BIBforeignlanguage{en-US}{2019 IEEE Symposium on Security and Privacy (SP)}}, May 2019. [Online]. Available: \url{http://dx.doi.org/10.1109/sp.2019.00031}
\BIBentrySTDinterwordspacing

\bibitem{Krizhevsky_2009}
A.~Krizhevsky, ``\BIBforeignlanguage{en-US}{Learning multiple layers of features from tiny images},'' Jan 2009.

\bibitem{Stallkamp_Schlipsing_Salmen_Igel_2011}
\BIBentryALTinterwordspacing
J.~Stallkamp, M.~Schlipsing, J.~Salmen, and C.~Igel, ``\BIBforeignlanguage{en-US}{The german traffic sign recognition benchmark: A multi-class classification competition},'' in \emph{\BIBforeignlanguage{en-US}{The 2011 International Joint Conference on Neural Networks}}, Jul 2011. [Online]. Available: \url{http://dx.doi.org/10.1109/ijcnn.2011.6033395}
\BIBentrySTDinterwordspacing

\bibitem{deng2009imagenet}
J.~Deng, W.~Dong, R.~Socher, L.-J. Li, K.~Li, and L.~Fei-Fei, ``Imagenet: A large-scale hierarchical image database,'' in \emph{2009 IEEE conference on computer vision and pattern recognition}.\hskip 1em plus 0.5em minus 0.4em\relax Ieee, 2009, pp. 248--255.

\bibitem{he2016deep}
K.~He, X.~Zhang, S.~Ren, and J.~Sun, ``Deep residual learning for image recognition,'' in \emph{Proceedings of the IEEE conference on computer vision and pattern recognition}, 2016, pp. 770--778.

\bibitem{huang2017densely}
G.~Huang, Z.~Liu, L.~Van Der~Maaten, and K.~Q. Weinberger, ``Densely connected convolutional networks,'' in \emph{Proceedings of the IEEE conference on computer vision and pattern recognition}, 2017, pp. 4700--4708.

\bibitem{simonyan2014very}
K.~Simonyan and A.~Zisserman, ``Very deep convolutional networks for large-scale image recognition,'' \emph{arXiv preprint arXiv:1409.1556}, 2014.

\bibitem{ioffe2015batch}
S.~Ioffe and C.~Szegedy, ``Batch normalization: Accelerating deep network training by reducing internal covariate shift,'' in \emph{International conference on machine learning}.\hskip 1em plus 0.5em minus 0.4em\relax pmlr, 2015, pp. 448--456.

\bibitem{ruder2016overview}
S.~Ruder, ``An overview of gradient descent optimization algorithms,'' \emph{arXiv preprint arXiv:1609.04747}, 2016.

\bibitem{Feng_Ma_Zhang_Zhao_Xia_Tao}
Y.~Feng, B.~Ma, J.~Zhang, S.~Zhao, Y.~Xia, and D.~Tao, ``\BIBforeignlanguage{en-US}{Fiba: Frequency-injection based backdoor attack in medical image analysis}.''

\bibitem{Chen_Liu_Li_Lu_Berkeley_Hannigan}
X.~Chen, C.~Liu, B.~Li, K.~Lu, U.~Berkeley, and A.~Hannigan, ``\BIBforeignlanguage{en-US}{Targeted backdoor attacks on deep learning systems using data poisoning}.''

\bibitem{Baluja_2017}
S.~Baluja, ``\BIBforeignlanguage{en-US}{Hiding images in plain sight: deep steganography},'' \emph{\BIBforeignlanguage{en-US}{Neural Information Processing Systems,Neural Information Processing Systems}}, Dec 2017.

\bibitem{barni2019new}
M.~Barni, K.~Kallas, and B.~Tondi, ``A new backdoor attack in cnns by training set corruption without label poisoning,'' in \emph{2019 IEEE International Conference on Image Processing (ICIP)}.\hskip 1em plus 0.5em minus 0.4em\relax IEEE, 2019, pp. 101--105.

\bibitem{Hu_Shen_Albanie_Sun_Wu_2020}
\BIBentryALTinterwordspacing
J.~Hu, L.~Shen, S.~Albanie, G.~Sun, and E.~Wu, ``\BIBforeignlanguage{en-US}{Squeeze-and-excitation networks},'' \emph{\BIBforeignlanguage{en-US}{IEEE Transactions on Pattern Analysis and Machine Intelligence}}, p. 2011–2023, Aug 2020. [Online]. Available: \url{http://dx.doi.org/10.1109/tpami.2019.2913372}
\BIBentrySTDinterwordspacing

\end{thebibliography}

\end{document}